\documentclass[10pt]{article}
\usepackage[margin=0.9in]{geometry}
\usepackage[numbers,sort&compress]{natbib}

\usepackage[utf8]{inputenc}
\usepackage[T1]{fontenc}
\usepackage[hidelinks]{hyperref}
\usepackage{url}
\usepackage{booktabs}
\usepackage{amsmath}
\usepackage{amsfonts}
\usepackage{amssymb}
\usepackage{nicefrac}
\usepackage{microtype}
\usepackage{xcolor}
\usepackage{graphicx}
\usepackage{multirow}
\usepackage{caption}
\date{}

\title{When Does Pre-Trained Capacity Help Causal Learning?\\From Scalar Prediction to Counterfactual Post-Training\thanks{Accepted at the NeurIPS 2026 Workshop: Transitioning from Pre-Training to Post-Training.}}
\author{%
  Qi Xu\thanks{Corresponding author: \texttt{xuqi0511@gmail.com}.}
  \quad Zijie Lei \quad Zhifei Deng \quad Zhigang Hua \quad Luke Simon \\
  Meta Platforms, Inc. \\
  Sunnyvale, CA, USA \\
}

\begin{document}
\maketitle
\begin{abstract}
How does pre-trained model capacity relate to performance after adaptation for causal tasks? We study two systems built on the Qwen 2.5 family: CausalLLM, which estimates scalar potential outcomes with treatment-specific heads, and CausalGen, which generates treatment-conditioned text or code using soft prefixes and factual supervised fine-tuning (SFT), counterfactual SFT, and preference alignment. Increasing backbone size from 0.5B to 1.5B yields mixed scalar-estimation results on IHDP, but improves CodeGen pass@1 from 0.267 to 0.575 and treatment-effect score from 0.231 to 0.515 after the complete generative pipeline. Within 0.5B CausalLLM, narrative inputs reduce mean $\sqrt{\mathrm{PEHE}}$ from 1.842 to 1.780, a 3.4\% improvement over tabular inputs; the 5.8\% improvement over a no-LLM baseline combines changes in representation and input format. These descriptive comparisons motivate a distinction between available and convertible capacity. They do not isolate the contribution of post-training: the two systems differ in dataset, architecture, output modality, and supervision, and intermediate-stage evaluations are unavailable. With only two small backbones and three replications, the results support a hypothesis about capacity--adaptation alignment rather than a general scaling law or a demonstrated causal mechanism.
\end{abstract}

\section{Introduction}
Modern language models are not produced by a single training stage. Pre-training establishes broad representations and generative capabilities, while supervised fine-tuning, preference optimization, and reinforcement-learning-style objectives are used to elicit and reshape downstream behavior. A central question is therefore not simply whether post-training improves a benchmark, but \emph{which capabilities are already present after pre-training, which are unlocked by task adaptation, and how model scale changes the effectiveness of the transition}.

Causal learning provides a useful testbed for this question because closely related tasks can require very different forms of adaptation. Classical individual treatment effect (ITE) estimation asks a model to predict scalar potential outcomes $Y(0)$ and $Y(1)$. Generative counterfactual modeling asks the richer question of producing the full outcome under an alternative intervention. Both require treatment-conditioned reasoning, but the former primarily tests whether pre-trained representations can be adapted into a causally useful predictive representation, whereas the latter additionally relies on the generative capabilities of the base model and on post-training procedures that teach the model how to express intervention-specific outcomes.

\begin{figure*}[t]
  \centering
  \includegraphics[width=\textwidth]{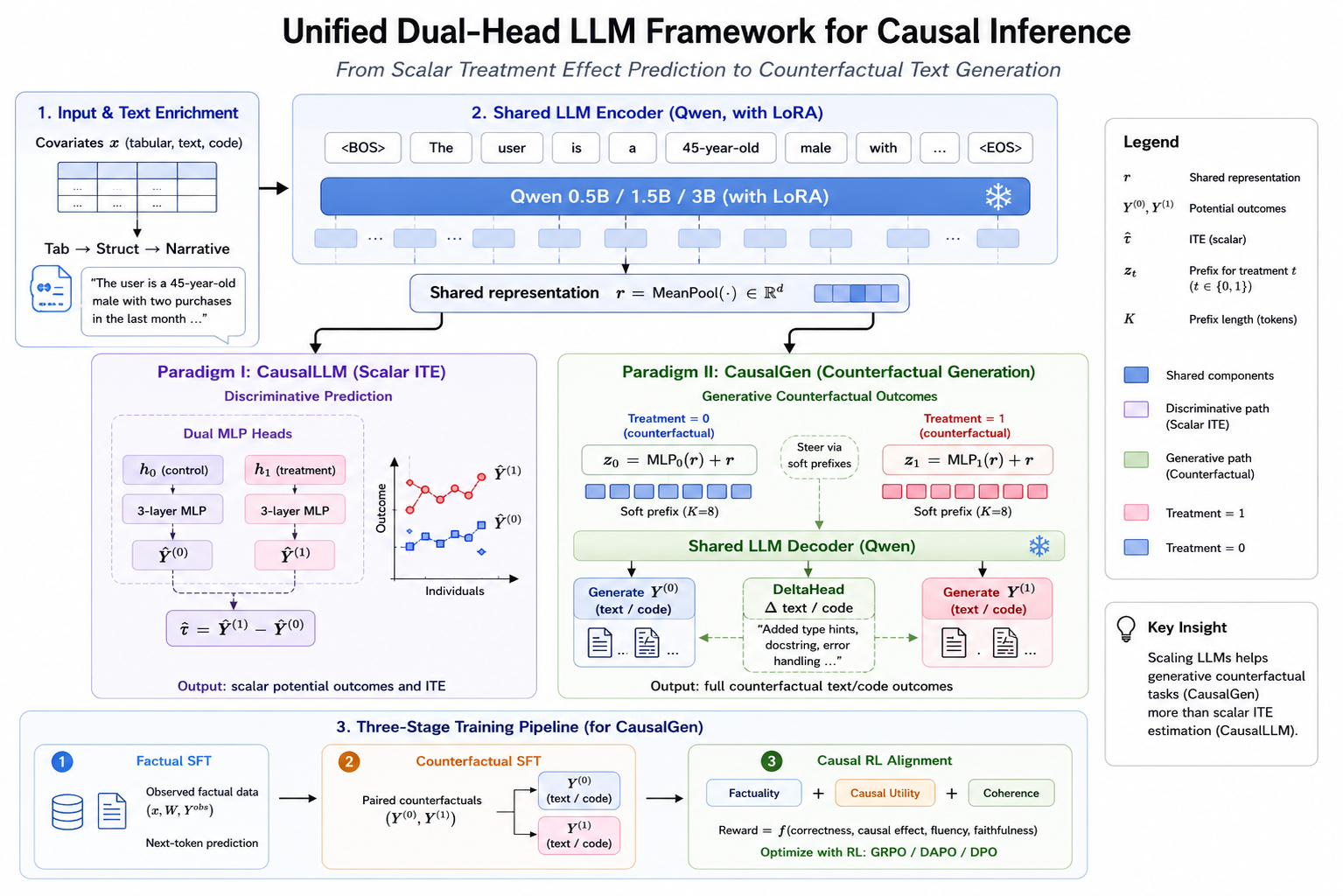}
  \caption{Two adaptation regimes built on pre-trained LLMs. (a) CausalLLM adapts a shared encoder with lightweight treatment-specific scalar heads for ITE estimation. (b) CausalGen uses treatment-specific projections and soft prefixes followed by factual SFT, counterfactual SFT, and causal preference alignment. Only the 0.5B and 1.5B backbones and DPO-style preference alignment are evaluated in this paper; the additional 3B, GRPO, and DAPO options shown in the schematic are not evaluated. The systems differ in data, heads, output modality, supervision, and training procedure; their scaling contrast is descriptive.}
  \label{fig:arch}
\end{figure*}

We study these regimes with two paradigms built on a common LLM family (Figure~\ref{fig:arch}). \textbf{CausalLLM} adapts a pre-trained LLM representation with treatment-specific scalar heads for $\hat{Y}(0)$ and $\hat{Y}(1)$; we additionally vary how structured covariates are rendered to test whether semantically meaningful inputs better expose knowledge acquired during pre-training. \textbf{CausalGen} replaces scalar heads with treatment-specific projections and soft prefixes that steer a shared decoder, and then applies a three-stage post-training pipeline: factual supervised fine-tuning (SFT), counterfactual SFT, and causal preference alignment.

The resulting comparison reveals different scaling patterns in the two final systems. On IHDP, $\sqrt{\mathrm{PEHE}}$ changes from 1.780 to 1.796, ATE error from 0.475 to 0.445, and RMSE from 1.247 to 1.259. On CodeGen, pass@1 and treatment-effect score increase by 2.15$\times$ and 2.23$\times$, respectively. Separately, narrative inputs improve mean $\sqrt{\mathrm{PEHE}}$ by 3.4\% relative to tabular CausalLLM. These observations motivate further controlled experiments; they do not establish that preference alignment or counterfactual SFT causes the larger model's advantage.

Our contributions are threefold: (1) we describe two related adaptation designs for scalar potential-outcome prediction and intervention-conditioned generation; (2) we report input-format and two-size comparisons within these systems; and (3) we formulate the distinction between \emph{available capacity} and \emph{convertible capacity} as a hypothesis for future capacity--adaptation studies. The common model family supports within-system comparisons, but the cross-system contrast is not a matched experiment because dataset, architecture, output modality, supervision, and training procedure change together.

\section{Related work}
\textbf{Deep treatment-effect estimation.} TARNet~\cite{shalit2017} learns shared representations with treatment-specific heads and balanced representation objectives; DragonNet~\cite{shi2019} augments this setup with propensity modeling and targeted regularization; TransTEE~\cite{zhang2022} uses transformer-based treatment--covariate interactions. These methods primarily evaluate scalar potential outcomes and treatment effects.

\textbf{LLMs and causal inference.} Kiciman et al.~\cite{kiciman2023} examine language models as sources of causal knowledge and reasoning. CLadder~\cite{jin2023} evaluates formal associational, interventional, and counterfactual queries, whereas CausalEval~\cite{causaleval2025} organizes and evaluates approaches that use language models as reasoning engines or as helpers for traditional causal methods. Liu et al.~\cite{liu2025} survey this broader interaction, including effect estimation. Our scalar estimator belongs to representation-based adaptation, while our generator evaluates intervention-conditioned outputs; neither setting alone establishes general causal reasoning ability.

\textbf{Post-training for causal tasks.} CausalGym~\cite{chen2026} directly compares SFT, DPO, KTO, PPO, and GRPO for causal inference tasks. This is closely related to our motivating question, but differs from our treatment-specific scalar heads and soft-prefix generation design. Unlike that systematic comparison of post-training approaches, our experiments evaluate CausalGen only after the complete pipeline. We therefore do not claim to isolate which post-training objective is responsible for its performance.

\textbf{Counterfactual language generation.} CausaLM~\cite{feder2021} uses counterfactual language-model representations for causal explanation, and Polyjuice~\cite{wu2021} produces controlled text perturbations. Bynum and Cho~\cite{bynum2025} use language-model-defined mechanisms in sequence-driven structural causal models to sample observational, interventional, and counterfactual data. CausalGen instead learns treatment-specific output generators from paired targets. Such targets supply supervision for generation, but generated plausibility or reference similarity does not by itself identify real-world causal effects.

\section{Method}

\subsection{CausalLLM: Scalar ITE}

A pre-trained LLM (Qwen 2.5-0.5B) encodes covariates via mean pooling: $\mathbf{r} = \text{MeanPool}(\text{LLM}(\text{tokenize}(\mathbf{x}))) \in \mathbb{R}^{896}$. Two 3-layer MLP heads ($896 \to 256 \to 128 \to 1$, ReLU, LayerNorm, Dropout) predict $\hat{y}(t) = h_t(\mathbf{r})$, with ITE $\hat{\tau} = h_1(\mathbf{r}) - h_0(\mathbf{r})$. Training uses factual loss with treatment routing:
\begin{equation}
\mathcal{L} = \tfrac{1}{n}\textstyle\sum_i \left[T_i \ell(h_1(\mathbf{r}_i), y_i) + (1{-}T_i) \ell(h_0(\mathbf{r}_i), y_i)\right]
\end{equation}
with optional MMD regularization ($\lambda_{\text{MMD}}{=}0.1$, Gaussian kernel with median heuristic bandwidth) to encourage balanced representations between treated and control groups. The LLM backbone is fine-tuned via LoRA~\cite{hu2022} (rank 16, $\alpha{=}32$, ${\sim}$0.5\% trainable parameters), preserving pre-trained knowledge while enabling task-specific adaptation.

\textbf{Text enrichment.} A key design choice is how tabular covariates are presented to the LLM. We compare three formatting modes of increasing semantic richness: \emph{Tabular} (raw numbers: ``0.5 1.0 0.0...''), \emph{Structured} (key-value pairs: ``age: 25, education: college, friends: 342''), and \emph{Narrative} (natural language: ``A 25-year-old college-educated user with 342 friends and moderate engagement...''). The hypothesis, not directly tested by representation probes or a randomly initialized backbone control, is that narrative mode better exposes the LLM's pre-trained understanding of real-world feature relationships---e.g., that education level correlates with employment outcomes---while tabular mode treats the LLM as a generic feature extractor.

\noindent\textbf{Treatment-conditioned alternative.} A natural treatment-conditioned alternative
is to concatenate a treatment token to the input and
evaluate the same model twice, once with $t=0$ and once with $t=1$,
yielding $\hat{\tau}(x)=f(x,1)-f(x,0)$. This single-head design evaluates the encoder separately for each treatment. Our dual-head formulation shares the covariate encoding and applies lightweight treatment-specific heads or prefixes. For scalar prediction, both outcomes can be computed from one encoding. For generation, producing both outcomes still requires two autoregressive decoding sequences; sharing the encoding does not eliminate this cost, and we report no measured latency advantage.

\subsection{CausalGen: Counterfactual Generation}

CausalGen shares the LLM encoder with CausalLLM but replaces MLP prediction heads with \textbf{dual projection heads} that transform the shared representation into treatment-specific representations via residual projection: $\mathbf{z}_t = \text{MLP}_t(\mathbf{r}) + \mathbf{r}$, where each MLP is a 2-layer network ($d_{\text{model}} \to 2d_{\text{model}} \to d_{\text{model}}$) with GELU activation. Each $\mathbf{z}_t$ is expanded into $K{=}8$ soft prefix tokens via learned linear projection: $\mathbf{P}_t = \text{Reshape}(W_{\text{prefix}} \mathbf{z}_t) \in \mathbb{R}^{K \times d_{\text{model}}}$. These prefix tokens are prepended to the target sequence embeddings, steering a \emph{shared} LLM decoder to generate $Y(t)$ as text or code---analogous to prefix tuning~\cite{li2021} but conditioned on causal representations. An optional \textbf{DeltaHead} generates treatment effect descriptions by composing $\mathbf{z}_0$ and $\mathbf{z}_1$ (via gating, cross-attention, subtraction, or bilinear modes).

Training uses three stages with different supervision. These stages describe the implemented pipeline; their individual necessity and incremental benefit are not established by the reported experiments:

\textbf{Stage 1: Factual SFT} (15 epochs). Each head is trained on observed outcomes only, with treatment routing: samples with $T{=}0$ train $h_0$, samples with $T{=}1$ train $h_1$. This establishes factual generation quality.

\textbf{Stage 2: Counterfactual SFT} (15 epochs). Both heads are trained jointly on paired $(Y(0), Y(1))$ data, forcing each head to generate the correct potential outcome: $\mathcal{L}_2 = \frac{1}{2}[\mathcal{L}_{\text{CE}}(h_0, Y(0)) + \mathcal{L}_{\text{CE}}(h_1, Y(1))]$.

\textbf{Stage 3: Causal preference alignment} (5 epochs). Preference optimization targets treatment-specific fidelity using treatment-derived preference pairs: for $h_1$, $Y(1)$ is ``preferred'' and $Y(0)$ is ``rejected''; vice versa for $h_0$. We use DPO-style preference optimization~\cite{rafailov2023}. GRPO and DAPO are not evaluated here. Opposite-treatment outcomes serve as rejected targets only when the intervention changes the outcome; unchanged or equally valid outcomes require careful pair filtering. Appendix~\ref{app:details} explains the conditioning and remaining implementation details.

\section{Experiments}

\textbf{Datasets.} CausalLLM: \textbf{IHDP}~\cite{hill2011} (747 samples, 25 covariates, semi-synthetic outcomes), \textbf{Twins}~\cite{louizos2017} (5K twin pairs, 15 covariates, binary mortality outcome). CausalGen: \textbf{CodeGen} (340 paired code samples with quality degradation treatment), \textbf{Jobs}~\cite{lalonde1986} (3.2K samples, job training treatment).

\textbf{Baselines.} CausalLLM: T-Learner (separate models), TARNet~\cite{shalit2017} (shared encoder + dual heads). CausalGen: single-head (one head for both outcomes), treatment-conditioned (treatment label in prompt).

\textbf{Implementation.} Unless otherwise stated, the main experiments use Qwen 2.5-0.5B with LoRA (rank 16, $\alpha{=}32$); the capacity study in Section~\ref{sec:scaling-contrast} repeats the matched setup with Qwen 2.5-1.5B. MLP baselines: Adam, lr=$10^{-3}$, 100 epochs. LLM models: AdamW, backbone lr=$2{\times}10^{-5}$, head lr=$10^{-3}$. CausalLLM: 20 epochs. CausalGen: 15+15+5 epochs across three stages. All results are reported as averages over 3 replications. Only Table~\ref{tab:causalllm} provides standard deviations for PEHE; the remaining summaries do not establish statistical significance. Appendix~\ref{app:details} consolidates the documented configuration and definitions.

\subsection{CausalLLM Results}

\begin{table}[t]
\caption{Scalar ITE estimation (3 reps, mean$\pm$std). ``Dual-head MLP'' is the no-LLM baseline shown as ``MLP (no LLM)'' in Fig.~\ref{fig:text}; CausalLLM (Narrative) uses the pre-trained LLM representation with the same treatment-specific prediction structure. $\downarrow$ = lower is better.}
\label{tab:causalllm}
\small
\begin{tabular*}{\textwidth}{@{\extracolsep{\fill}}lcccc@{}}
\toprule
& \multicolumn{2}{c}{\textbf{IHDP}} & \multicolumn{2}{c}{\textbf{Twins}} \\
\cmidrule(lr){2-3}\cmidrule(lr){4-5}
\textbf{Method} & $\sqrt{\text{PEHE}}$$\downarrow$ & ATE error$\downarrow$ & $\sqrt{\text{PEHE}}$$\downarrow$ & ATE error$\downarrow$ \\
\midrule
T-Learner & 1.96{\tiny$\pm$.05} & \underline{0.21} & .467{\tiny$\pm$.012} & .056 \\
TARNet & 2.02{\tiny$\pm$.08} & 0.34 & .454{\tiny$\pm$.012} & .012 \\
\midrule
Dual-head MLP (no LLM) & 1.89{\tiny$\pm$\textbf{.03}} & 0.31 & .452{\tiny$\pm$\textbf{.007}} & .035 \\
CausalLLM (Narrative) & \underline{1.78}{\tiny$\pm$.19} & 0.48 & \underline{.450}{\tiny$\pm$.009} & \underline{.012} \\
\bottomrule
\end{tabular*}
\end{table}

\begin{figure}[t]
  \centering
  \includegraphics[width=\linewidth]{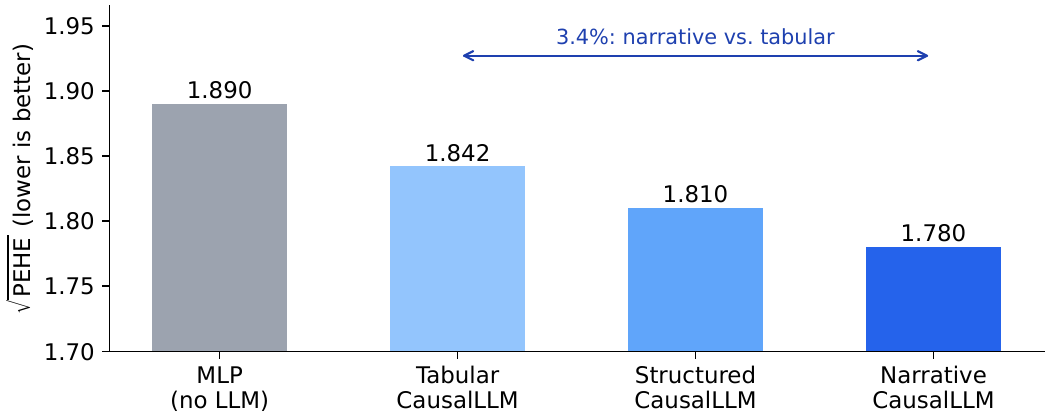}
  \caption{Input-format comparison (IHDP, 0.5B). Narrative inputs reduce mean $\sqrt{\mathrm{PEHE}}$ by 3.4\% versus tabular CausalLLM. The 5.8\% reduction versus the no-LLM MLP is a different, combined comparison; uncertainty for the format differences is not available.}
  \label{fig:text}
\end{figure}

On IHDP, narrative CausalLLM has lower mean $\sqrt{\mathrm{PEHE}}$ than the no-LLM dual-head MLP (1.780 versus 1.890), a 5.8\% relative reduction. This comparison changes both the representation and input rendering and cannot isolate either contribution. Within CausalLLM, the mean errors for tabular, structured, and narrative inputs are 1.842, 1.810, and 1.780, respectively (Figure~\ref{fig:text}). The narrative--tabular reduction is $(1.842-1.780)/1.842=3.4\%$, while structured--tabular and narrative--structured reductions are 1.7\% and 1.7\%. These mean differences are descriptive: three replications and the reported IHDP variability do not justify a significance claim.

The metric tradeoff also matters. On IHDP, narrative CausalLLM's ATE error (0.48) is worse than each listed baseline (0.21, 0.34, and 0.31), despite its lower mean PEHE. On Twins, its PEHE is numerically close to the no-LLM baseline (0.450 versus 0.452), and its rounded ATE error ties TARNet (0.012). Thus, the evidence does not support a uniform advantage across datasets or estimands.

\subsection{CausalGen Results}

\begin{table}[t]
\caption{Counterfactual generation (3 reps). The final dual-head system has 2.3--2.5$\times$ higher reference BLEU for treatment arm $Y(0)$.}
\label{tab:causalgen}
\small
\begin{tabular*}{\textwidth}{@{\extracolsep{\fill}}lcccc@{}}
\toprule
& \multicolumn{2}{c}{\textbf{CodeGen}} & \multicolumn{2}{c}{\textbf{Jobs}} \\
\cmidrule(lr){2-3}\cmidrule(lr){4-5}
\textbf{Method} & BLEU $Y(0)$ & TE Score & BLEU $Y(0)$ & TE Score \\
\midrule
Single-head & 0.151 & --- & 0.193 & --- \\
\midrule
CausalGen & \textbf{0.374} & \textbf{0.354} & \textbf{0.440} & 0.312 \\
\quad $\Delta$ vs.\ single & \textbf{2.5$\times$} & & \textbf{2.3$\times$} & \\
\bottomrule
\end{tabular*}
\end{table}

CausalGen obtains 2.5$\times$ higher reference BLEU for $Y(0)$ on CodeGen (0.374 versus 0.151) and 2.3$\times$ on Jobs (0.440 versus 0.193). These comparisons favor the final dual-head system, but do not separately identify architecture and training-stage effects. Reference similarity is also not equivalent to functional correctness or causal identification. Although the original experiment description labels $Y(0)$ counterfactual and $Y(1)$ factual, this designation is treatment-dependent: for an example with observed $T_i$, the factual outcome is $Y_i(T_i)$ and the counterfactual is $Y_i(1-T_i)$. Interpretation of the arm-specific scores therefore requires the evaluation cohort's treatment assignment and paired-target construction.

\smallskip
\noindent\textbf{Generation example.} On CodeGen, given
\texttt{def calculate\_area(shape, dims):}, CausalGen generates
treatment-specific counterfactual implementations. $Y^{(0)}$ (without
code review) contains a minimal implementation with hardcoded constants
and limited validation, while $Y^{(1)}$ (with code review) adds type
hints, docstrings, input validation, and replaces hardcoded values with
\texttt{math.pi}. The DeltaHead summarizes the inferred treatment effect
as: ``Added documentation, validation, and numerical precision
improvements.'' On Jobs, given age 28, high-school education, and prior
earnings of \$4,200, $Y^{(0)}$ describes prolonged unemployment and
unstable work, while $Y^{(1)}$ describes obtaining stable employment
after training.

\subsection{DeltaHead Ablation}

The DeltaHead composes $\mathbf{z}_0$ and $\mathbf{z}_1$ to generate treatment effect text. We compare four composition modes on CodeGen (Table~\ref{tab:deltahead}).

\begin{table}[t]
\caption{DeltaHead ablation (CodeGen, 0.5B, 3 reps).}
\label{tab:deltahead}
\small
\begin{tabular*}{\textwidth}{@{\extracolsep{\fill}}lcccc@{}}
\toprule
\textbf{Mode} & \textbf{pass@1} & \textbf{TE Score} & \textbf{CF Faith.} & \textbf{Swap} \\
\midrule
Subtract & 0.100 & 0.294 & 0.373 & 0.611 \\
Gate & 0.167 & 0.354 & \textbf{0.429} & 0.592 \\
Cross-Attn & \textbf{0.408} & \textbf{0.363} & 0.379 & \textbf{0.654} \\
Bilinear & 0.200 & 0.213 & 0.377 & 0.430 \\
\bottomrule
\end{tabular*}
\end{table}

Cross-attention has the highest reported pass@1 (0.408), TE Score (0.363), and Swap score (0.654). Gate has the highest CF Faithfulness (0.429), using adaptive gating over $[\mathbf{z}_1;\mathbf{z}_0;\mathbf{z}_1-\mathbf{z}_0;\mathbf{z}_1\odot\mathbf{z}_0]$. Thus, the table shows a tradeoff rather than a single best composition. Without uncertainty estimates, the difference between cross-attention and gating TE scores (0.009) should not be treated as statistically established. The generation and scaling tables represent separate reported settings; their 0.5B rows are not identical and cannot be pooled as a single configuration.

\subsection{A Task-Dependent Capacity Effect}\label{sec:scaling-contrast}

\begin{table}[t]
\caption{Effect of increasing backbone size from 0.5B to 1.5B. Scalar ITE estimation on IHDP shows no consistent gain across error metrics, whereas counterfactual generation on CodeGen improves substantially. We treat this two-size comparison as a task-dependent capacity effect, not a scaling law.}
\label{tab:scaling}
\small
\begin{tabular*}{\textwidth}{@{\extracolsep{\fill}}lccc|ccc@{}}
\toprule
& \multicolumn{3}{c|}{\textbf{CausalLLM (IHDP)}} & \multicolumn{3}{c}{\textbf{CausalGen (CodeGen)}} \\
\cmidrule(lr){2-4}\cmidrule(lr){5-7}
\textbf{Size} & $\sqrt{\text{PEHE}}$ & ATE err. & RMSE & pass@1 & TE & ATE dir.\% \\
\midrule
0.5B & \textbf{1.780} & 0.475 & \textbf{1.247} & 0.267 & 0.231 & 73.3 \\
1.5B & 1.796 & \textbf{0.445} & 1.259 & \textbf{0.575} & \textbf{0.515} & \textbf{93.3} \\
\bottomrule
\end{tabular*}
\end{table}

Table~\ref{tab:scaling} shows a task-dependent response to additional pre-trained capacity. Moving from Qwen 2.5-0.5B to 1.5B does not consistently improve scalar ITE estimation on IHDP: $\sqrt{\mathrm{PEHE}}$ changes from 1.780 to 1.796, ATE error improves from 0.475 to 0.445, and RMSE changes from 1.247 to 1.259. In contrast, the same increase in backbone size produces a large gain for counterfactual generation on CodeGen.

\textbf{Scalar ITE shows no consistent capacity benefit in this comparison.} The mixed movement across $\sqrt{\mathrm{PEHE}}$, ATE error, and RMSE indicates that a larger pre-trained backbone alone is not sufficient to improve scalar causal estimation in our setting. One possible explanation is that scalar ITE performance is constrained by causal supervision and treatment-specific adaptation rather than generic semantic capacity, but our experiments do not isolate the mechanism.

\textbf{Counterfactual generation benefits strongly from the larger backbone.} From 0.5B to 1.5B, pass@1 rises 2.15$\times$ (0.267$\to$0.575), TE Score rises 2.23$\times$ (0.231$\to$0.515), and ATE-direction accuracy rises from 73.3\% to 93.3\%. These results describe better final-system generation metrics under the reported CodeGen pipeline. They do not distinguish improvements already present in the larger base model from those attributable to individual adaptation stages.

\textbf{Interpretation.} The contrast motivates a distinction between \emph{available capacity} (the backbone's parameter and representation resources) and \emph{convertible capacity} (the portion that an adaptation setup can use for a target task). These are conceptual terms, not independently measured quantities. Counterfactual SFT and preference alignment are plausible pathways for exploiting capacity, but the present experiments cannot attribute the scaling difference to them. Dataset, architecture, output modality, and supervision all differ across systems.

\section{Discussion and Limitations}
\textbf{What the experiments establish.} We observe mixed 0.5B-to-1.5B scalar metrics on IHDP, improved final-system generation metrics on CodeGen, and lower mean PEHE with narrative rather than tabular inputs within 0.5B CausalLLM. These results suggest that task design and input representation deserve attention alongside parameter count. They do not demonstrate a universal relation between pre-training scale and causal capability.

\textbf{What remains unisolated.} We do not evaluate an untuned base model or intermediate checkpoints after factual SFT and counterfactual SFT at both sizes. Consequently, the marginal contributions of paired supervision and preference alignment, and their interaction with capacity, are unknown. Likewise, the input-format study lacks additional replications and controls for semantic content, sequence length, and random initialization; it cannot establish that narrative rendering activates particular pre-trained knowledge.

\textbf{Scope and statistical uncertainty.} Only two small Qwen 2.5 backbones are compared, and the datasets are small, especially CodeGen (340 paired examples before splitting). Three replications and missing dispersion for most generation metrics limit statistical confidence. The cross-system comparison changes dataset, architecture, output type, supervision, and training procedure together. Neither larger-model behavior nor transfer across model families is established. The IHDP ATE tradeoff further prevents a claim of uniformly better causal estimation.

\textbf{Causal identification and target provenance.} Potential-outcome estimation from observational data relies on consistency, overlap, and conditional exchangeability; an LLM representation or balancing penalty does not guarantee these assumptions. Paired generation targets provide supervised alternative outcomes, but their validity depends on their construction. Jobs does not ordinarily contain both individual potential outcomes, so reference narratives must not be interpreted as observed individual counterfactual truth. The construction and evaluation of those paired targets require explicit documentation. BLEU and qualitative examples measure output agreement or illustrate behavior, rather than proving real-world causal validity.

\textbf{Next controlled tests.} A stronger study would evaluate the base model and every post-training checkpoint at both sizes on the same held-out examples; train scalar and generative outputs on one paired dataset with identical information and splits; and add seeds, confidence intervals, larger backbones, and another model family. Validation-selected checkpoints are also needed to assess whether the 100-epoch MLP schedule overfits. These tests are proposed work, not completed experiments.

\section{Conclusion}
Our two adaptation systems exhibit different reported capacity responses: scalar potential-outcome estimation has mixed metric changes, whereas the fully adapted counterfactual generator improves at 1.5B. Narrative inputs also reduce mean scalar PEHE relative to tabular inputs, with uncertain statistical support. These observations motivate available versus convertible capacity as a useful hypothesis. Establishing the mechanism requires matched data and stage-by-stage evaluations that separate pre-training, architecture, supervision, and optimization.

\clearpage
\appendix
\section{Experimental Details and Definitions}\label{app:details}
This appendix consolidates the documented configuration and distinguishes standard definitions from implementation details not available in the experiment record.

\subsection{Scalar estimation}
For covariates $x_i$ and binary treatment $T_i$, the observed outcome is $Y_i(T_i)$ and the individual treatment effect is $\tau_i=Y_i(1)-Y_i(0)$. CausalLLM predicts $\hat\tau_i=\hat Y_i(1)-\hat Y_i(0)$. The standard scalar metrics are
\begin{align}
\sqrt{\mathrm{PEHE}}&=\sqrt{\frac{1}{n}\sum_i(\hat\tau_i-\tau_i)^2},\\
\mathrm{ATE\ error}&=\left|\frac{1}{n}\sum_i\hat\tau_i-\frac{1}{n}\sum_i\tau_i\right|.
\end{align}
ATE is an estimand, whereas the downward-arrow scalar-table entries refer to its absolute estimation error. Reference individual effects are required for PEHE. The standard factual-outcome RMSE is $\sqrt{n^{-1}\sum_i(\hat Y_i(T_i)-Y_i(T_i))^2}$; whether the reported RMSE uses factual outcomes or both potential outcomes remains to be verified against the evaluation code.

Factual training routes each unit only to its observed-treatment head. The optional MMD regularizer compares treated and control representations using a Gaussian kernel with median-heuristic bandwidth and weight 0.1. Which reported runs enable MMD is not specified. The input-format examples in the method are schematic, not the exact IHDP or Twins templates. A format comparison should retain identical pre-treatment information and exclude outcomes, post-treatment variables, or inferred facts from narratives.

\subsection{Soft prefixes and generation supervision}
The documented generator computes $z_t=r+\mathrm{MLP}_t(r)$ and $P_t=\mathrm{Reshape}(W_{\mathrm{prefix}}z_t)\in\mathbb R^{8\times d}$. Each prefix row is a continuous embedding, not a vocabulary token. Treatment selects the projection and prefix, which precede the target embeddings and condition a shared autoregressive decoder. Both requested outcomes require separate decoding sequences despite sharing the covariate encoding.

The sequence objective is cross-entropy on target outcomes. Factual SFT supervises only the observed arm, whereas counterfactual SFT supervises both paired targets with equal weight. A standard expression of the stated conditioning is
\begin{equation}
-\sum_{j=1}^{|Y(t)|}\log p_\theta\big(Y(t)_j\mid P_t,Y(t)_{<j}\big).
\end{equation}
This expresses the design rather than an exact training prompt. Additional covariate or instruction tokens, attention masks, position handling, truncation, and target-loss masking are not specified in the supplied experiment record.

Stage 3 uses $Y(t)$ as the preferred outcome and $Y(1-t)$ as the rejected outcome for arm $t$. Standard DPO regularizes a preference likelihood ratio relative to a reference policy~\cite{rafailov2023}. The implemented reference checkpoint, preference strength, pair filtering, and sequence normalization are not specified. Unchanged or equally valid outcomes need careful filtering. This offline preference objective does not imply online RL rollouts; no GRPO or DAPO comparison is reported.

\subsection{Baselines and hyperparameters}
The T-Learner fits independent treatment and control models, TARNet uses a shared representation and treatment-specific heads, and the dual-head MLP provides the no-LLM scalar comparison. The generation table reports a single-head baseline. The treatment-conditioned alternative is described but has no numerical result in that table. Exact baseline layer sizes, regularization, and whether generator baselines receive identical training data and stages are not specified, so the comparisons do not isolate architecture alone.

\begin{table}[ht]
\centering\small
\caption{Documented experimental configuration.}
\begin{tabular}{@{}lp{0.64\textwidth}@{}}
\toprule
Setting & Value \\
\midrule
Backbone family & Qwen 2.5; 0.5B and 1.5B in the size comparison \\
LoRA & Rank 16; scaling $\alpha=32$ \\
Scalar heads (0.5B) & $896\to256\to128\to1$; ReLU, LayerNorm, dropout \\
Generative projection & $d\to2d\to d$; GELU; residual connection \\
Soft prefix & 8 continuous embeddings \\
MLP optimizer & Adam; learning rate $10^{-3}$; 100 epochs \\
LLM optimizer & AdamW; backbone rate $2\times10^{-5}$; head rate $10^{-3}$ \\
Scalar adaptation & 20 epochs \\
Generative stages & 15 factual + 15 paired SFT + 5 preference epochs \\
Repetitions & 3; PEHE standard deviations in the scalar table \\
\bottomrule
\end{tabular}
\end{table}

These settings do not establish validation-based hyperparameter selection. Batch size, dropout rate, weight decay, learning-rate schedule, LoRA target modules, checkpoint selection, seeds, and split sizes are not available. A 100-epoch MLP schedule alone establishes neither overfitting nor its absence; validation curves and checkpoint selection are needed. The experiment record also does not explain the different 0.5B generator scores in the composition and size studies.

\subsection{Generation metrics and target provenance}
BLEU measures reference overlap rather than causal validity. Reproduction requires tokenization, n-gram order, smoothing, aggregation, and code normalization. In code generation, standard pass@1 measures the probability that one candidate passes the task's tests. The actual tests, decoding settings, evaluation denominator, and use of that standard protocol need confirmation. TE Score, CF Faithfulness, Swap, and generation ATE-direction percentage are task-specific metrics; scoring functions, references, tie handling, and aggregation are not provided in the experiment record. These quantities cannot be equated with scalar PEHE or ATE error.

The described generation datasets contain 340 paired CodeGen examples and 3.2K Jobs examples. The code illustration uses code review, whereas the dataset description refers to quality degradation; their precise treatment mapping needs documentation. CodeGen's paired-target construction, split, and related-code leakage controls are not specified. Jobs ordinarily contains observed earnings rather than both individual potential outcomes. Its conversion to paired outcome narratives, and whether they are synthetic or model-generated, therefore needs documentation. Generated narratives should not be interpreted as observed individual counterfactual truth. The main-text examples illustrate behavior rather than providing this provenance.

\section{A Controlled Follow-up Design}
To test capacity--adaptation interactions, evaluate the untuned base, factual-SFT, counterfactual-SFT, and preference-aligned checkpoints for both sizes on identical held-out examples. Fix prompts, decoding, and evaluation. Stage-to-stage changes quantify incremental gains; differences in those changes across model sizes test whether a stage benefits the larger model more. Multiple seeds and paired uncertainty estimates are needed. No missing checkpoint results are imputed in this revision.

A shared paired dataset could provide complete outcomes and pre-specified scalar summaries, allowing both output tasks to use identical covariates, treatments, splits, and supervised information. Input-format controls should preserve information and record token lengths. Larger backbones and another family would assess external validity. These are proposed controls, not completed experiments.

\end{document}